\documentclass[preprint,12pt,authoryear]{elsarticle}
\makeatletter
\def\ps@pprintTitle{%
 \let\@oddhead\@empty
 \let\@evenhead\@empty
 \def\@oddfoot{}%
 \let\@evenfoot\@oddfoot}
\makeatother

\usepackage{amssymb}
\usepackage[ruled]{algorithm2e}
\usepackage{adjustbox}
\usepackage{tabularx}
\usepackage{graphicx}
\usepackage{booktabs}

\SetAlFnt{\small}
\SetAlCapFnt{\small}
\SetAlCapNameFnt{\small}
\SetAlCapHSkip{0pt}
\IncMargin{-\parindent}

\begin{document}

\begin{frontmatter}



\title{Leveraging Conversation Structure on Social Media \\to Identify Potentially Influential Users}

\author[label1]{Dario De Nart}
\ead{dario@datamantix.com}
\author[label1]{Dante Degl'Innocenti}
\ead{dante@datamantix.com}
\author[label1]{Marco Pavan}
\ead{marco@datamantix.com}
\address[label1]{Datamantix S.r.l., Viale Tricesimo 200, Udine, Italy\fnref{website}}

\fntext[website]{Datamantix S.r.l. website \texttt{\textbf{http://datamantix.com}}}

\begin{abstract}

Social networks have a community providing feedback on comments that allows to identify opinion leaders and users whose positions are unwelcome. Other platforms are not backed by such tools. Having a picture of the community’s reactions to a published content is a non trivial problem. In this work we propose a novel approach using Abstract Argumentation Frameworks and machine learning to describe interactions between users. Our experiments provide evidence that modelling the flow of a conversation with the primitives of AAF can support the identification of users who produce consistently appreciated content without modelling such content.

\end{abstract}


\end{frontmatter}


\section{Introduction}
\label{sec:intro}
\begin{figure}[ht]
    \centering
    \includegraphics[width=0.7\textwidth]{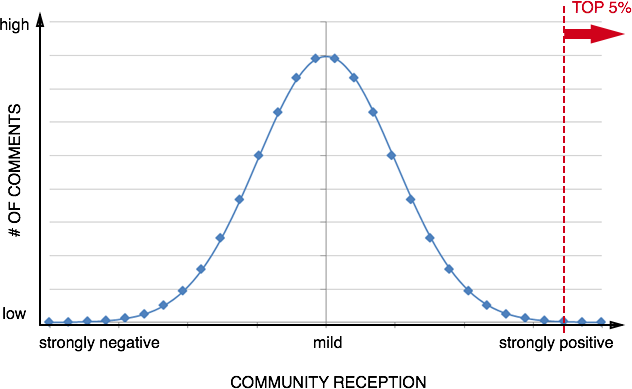}
    \caption{How we assume the posts to be distributed with respect to the community's reception}
    \label{fig:user_distr}
\end{figure}

Detecting opinion leaders and popular users within an online community is a desirable task for a number of practical applications such as social media activity monitoring, content placing, and trend detection.
All these tasks, if effectively carried out, can substantially lift the value of a community, however the detection of users who consistently generate content appreciated by the community is not always straightforward. 
We assume that a community's reception of user-generated content will be somewhat close to a Gaussian one, like is shown in Figure \ref{fig:user_distr}, with a relatively small number of items attracting the most negative feedback, another small fraction attracting the most positive feedback, and the vast majority receiving a mild reception or just little attention.
While most literature effort is directed towards describing \cite{buckels2014trolls,correa2010interacts} or detecting \cite{arnt2003learning}  the ``bad'' users or content that occupies the left hand end of the distribution, in this work we focus on detecting the ``good'' ones, standing on the right hand end.
This task is of particular interest since recent studies show evidence of how positive feedback tends to generate more herding effect in Social media than negative one \cite{muchnik2013social} and therefore can be regarded as more interesting from a community management point of view as well.
Some platforms provide their users with the possibility of expressing explicit feedback on user-generated content by means of upvoting and downvoting. The proportion between upvotes and downvotes is a simple and straightforward indicator of how the community received the published content: a strong majority of positive feedback clearly indicates a well received content and the identification of users publishing appreciated content in regular basis is rather straightforward.
However not all social media provide such tools or allow external applications to access them.
As a matter of fact most platforms, like Facebook and Twitter, allow only positive feedback to be explicitly expressed by means of voting rather than commenting.
Sentiment Analysis is a valuable tool to understand the reactions of a community to a given content \cite{pang2008opinion}, however Sentiment Analysis systems, being trained on the textual contents of a set of labelled texts, are highly conditioned by writing style nuances and the considered domain of application.
Moreover the usage of negation \cite{wiegand2010survey} and irony \cite{bosco2013developing} are notoriously hard problems to face when building a Sentiment Analysis application.

We claim that the very structure of the conversations hosted on Social Media platforms embeds information about the sympathy and overall appreciation aroused by a post, and hence the reputation of its author.
By leveraging such information, it is possible to abstract over the content of the conversation and therefore avoid non-trivial problems such as the aforementioned negation and irony managing.
To support our claim, we present in this work a novel approach exploiting Dung's well known Abstract Argumentation Framework in the context of social media. While other studies focus on understanding the topic of the argumentation, identifying the coalitions of arguments or users, or determining which are the strongest arguments, our work is content agnostic and aims at modelling the social interaction of users involved in an online conversation to extract from its flow insights about users' social influence.

The rest of the paper is organised as follows: in Section~\ref{sec:relWork} we introduce some related work, in Section~\ref{sec:conversationModelling} we provide a brief overview of Abstract Argumentation and describe the procedure we used to build a AAF from a social media conversation, in Section~\ref{sec:consideredFeatures} we describe the features extracted from the AAF and provide an overview of the used classifiers, in Section~\ref{sec:dataAnalysisEval} we present an evaluation of our technique performed over Reddit conversations. 
We discuss the results in Section~\ref{sec:discussion}, and we finally draw conclusions and describe some future works directions in Section~\ref{sec:conclusions}.

\section{Related Work}
\label{sec:relWork}
Since their birth, social network sites have attracted a growing interest in the research community and have been analysed by several authors in the literature. Extensive data mining activities have been performed to address a broad span of problems including text categorisation, user profiling, communities detection, and content recommendation \cite{abel11}. Noticeable examples of this kind research can be found in~\cite{social-enrich}, where the authors exploit contextual enrichment to improve the topic extraction process from texts posted on the Twitter platform; in~\cite{twitter-sim} a new user model approach has been proposed to compute user similarity based on a network representing the semantic relationships between the words occurring in the same tweet and the related topics; in~\cite{tao12} is presented a user model that features topic detection and entity extraction for tweets and links the latter to news articles to describe the context of the tweets; finally, a recent survey~\cite{riquelme15} illustrates how there is a growing interest in measuring user influence on Twitter.
Understanding the behavior of users on various social media platforms, their interaction paradigms \cite{kirschner2015facebook}, and predicting which contents are potentially harmful for the community are still open and debated problems. In \cite{arnt2003learning} a seminal methodology for automated moderation leveraging relatedness between posts and a Naive Bayes classifier is presented. In \cite{arora2015good} a content-based technique to predict which Stack Overflow answers will get negative scores is presented.
Most of the proposed approaches, however, rely on content-based features to assess the potential reception of the community to a given post, giving little importance or ignoring the discourse structure of social media interaction.

The problem of analysing discourse and in particular argumentation has recently become prominent in the Artificial Intelligence research community with several studies focusing on externalising implicit argumentation structures hidden in user generated content. The authors of \cite{rahwan2009argumentation} provide a comprehensive introduction to the study of argumentation in Artificial Intelligence, including theoretical and computational aspects of the subject.
Several formal frameworks have been proposed to describe user argumentation, mostly focused on claims and justifications. Ontological descriptions of argumentation like the SIOC argumentation ontology module \cite{lange2008expressing}, DILIGENT \cite{tempich2005argumentation}, and several others \cite{schneider2013review} have been developed, but found little application due to their intrinsic complexity.
Machine learning based classification of social media argumentation has been explored as well, however, as pointed out by the authors of \cite{llewellyn2014re} the scarce availability of human annotated trained corpora and the domain dependency of such approaches are severe limiting factors.
Abstract Argumentation Frameworks (herein AAF), on the other hand, are a powerful, yet simple graph based representation of argumentation, introduced in \cite{dung1993acceptability,dung1995acceptability}; the main advantage of Dung's AAF is the abstraction over the actual content of single arguments and the focus on the relationships among considered arguments, which is a substantial difference from more traditional approaches that leverage the breakdown of an argument into claim, premises, and evidence \cite{toulmin2003uses}.
AAF allow the evaluation of sets of acceptable arguments, may it be a binary labelling (accepted or rejected) or a ranking with several levels of acceptability. This latter approach can come in many flavours and a detailed comparative survey of ranking-based semantics for AAF is presented in \cite{bonzon2016comparative}. Over the years AAF found application in different domains, each of them requires specific semantics \cite{charwat2015methods} and specific extensions of the original attack-based model. The authors of \cite{leite2011social} investigate the usage of AAF in the context of social media, highlighting how due to the large number of arguments to be considered in such a scenario, ranking based approaches are to be preferred, moreover introduce a Social extension of AAF considering also the votes (such as likes, thumbs up, and so on) expressed by the community. The authors of \cite{grosse2015integrating} propose an extension of the AAF to mine opinions from Twitter.
Another substantial extension of AAF are Bipolar AAF \cite{cayrol2009bipolar,cayrol2010coalitions} that include the notion of support as a primitive. Such a representation can be further extended by including other relations among arguments, such as noncommittal, in an arbitrary high number \cite{brewka2013abstract}.
Recent developments in Natural Language Processing in particular allow the extraction of claims \cite{ecklekohler-kluge-gurevych:2015:EMNLP}, supporting evidence \cite{rinott2015show}, and other key components of argumentation with satisfactory precision.
The authors of \cite{lippi2016argumentation} provide a detailed survey of the state of the art in argument mining to which we address the curious reader.
However in this work we are not mining actual arguments out of our conversation, but rather exploiting AAF as a modelling framework to describe the flow of an online conversation in which several users interact, thus argumentation mining is out of our scope.

\section{Conversation Modelling}
\label{sec:conversationModelling}
In this section we present the modelling framework we designed to describe the interactions between users occurring within a conversation and the methodology we use to build such a model.
In the first part of this section we will briefly introduce the basic concepts of AAF, then we will pinpoint the assumption we introduced to fit AAF modelling to our problem, and finally we will illustrate the graph construction rules we used to model Social Media conversation. 
\subsection{Abstract Argumentation Basics}
As introduced in Section \ref{sec:relWork}, Abstract Argumentation Frameworks \cite{dung1993acceptability,dung1995acceptability} provide a simple, yet powerful, network representation of the structure of argumentation.
Formally speaking, an argumentation framework is a pair $(AR, attacks)$ where $AR$ is the set of considered arguments, and $attacks$ is a binary relation on $AR$.
Since the attack relationship is binary, a discourse can be represented as a directed graph.
The notion of \emph{attack} is the key concept of an abstract argumentation framework: arguments are meant to attack each other and the structure of the network of attacks between arguments embeds the semantic information we aim to extract.
The word attack in the context of AAF must not be misinterpreted since it does not imply aggressive behaviour, but rather represents an argument trying to weaken another one.

To further expand the representation of the interaction between users, supporting arguments can be modelled as well.
Intuitively, an argument can be considered supporting if it strengthens another one, either by bringing actual supporting evidence or by attacking its attacking arguments.
The adoption of support edges as well allows to model a discourse as a multigraph.

One of the main advantages of adopting AAF is their ability to introduce an abstraction layer over the contents of the discourse and the nature of the attacking and supporting arguments, hence, once built, AAF can be considered content agnostic which is coherent to our goal.

\subsection{Considerations and Assumptions on Social Media Discourse}
Abstract argumentation frameworks are typically used to detect \emph{conflict free} sets of arguments, i.e. arguments that do not attack each other and therefore are most likely to be accepted by the participants to the discussion\footnote{Being conflict-free however is a necessary but not sufficient condition for a set of arguments to be accepted: an argument set should also be admissible in that it defends all its members against attack. Moreover, there can be multiple and mutually conflicting admissible sets in the same discussion.}.
However, our point is not finding which arguments are considered valid by the community, but rather which posts and authors attract the most sympathy and appreciation.
To fit this goal, some assumptions must be relaxed and some others must be introduced.
First and foremost, reasoning over AAF implies assuming that the involved agents are \emph{rational}, which in the context of social media platforms is a strong assumption since emotional factors are prominent and greatly affect the interaction between users \cite{kramer2014experimental}.
Moreover, sympathy is not rational by definition, hence we will not consider our agents as rational and we will not perform actual reasoning on the constructed argument graphs.
On the other hand, we assume that the discourse structure provided by AAF modelling embeds behavioural intentions and therefore patterns implying sympathy and appreciation can be learnt from it.

The discourses hosted on Social Media platforms can arguably differ significantly from the ones conducted on more traditional media due to their peculiar characteristics and constraints.
We can pinpoint four main characteristics which are relevant to extents of our modelling activity.
\begin{itemize}
\item \emph{Brevity}: posts on a social media platform are typically short\footnote{Some platforms, like Twitter, force users to produce short texts.}; this fact makes unlikely to attack and to defend the same user within a single post's space.
\item \emph{No need of writing to express support}: typically Social Media platforms offer their users tools to express positive feedback (such as like, re-tweet, or similar options) that do not imply writing a reply. On the other hand, on most platforms, writing a comment is the only way to express criticism. 
\item \emph{Multiple posting}: it is not unusual for social media users to post sequences of consecutive messages and those self-replies are highly unlikely to be attacks. This behaviour can be seen as a consequence of the aforementioned typical brevity of social media posts.
\item \emph{Referencing}: it is generally easy to identify to whom a post is replying since most social platforms provide reply indentation\footnote{Plaforms such as Reddit and Youtube indent posts according to whom they reply to.} or an explicit annotation\footnote{Platforms such as Twitter and Facebook provide a referral to the replied post or to its author}, hence there is no need to infer it from the textual content of a post.
\end{itemize}
Attack and support relationships are defined in our approach, which exploits AAF concepts, leveraging the above considerations.
We define as an attack to a given post any replying message authored by a distinct user from the one that wrote the original message.
We consider messages including explicit referrals to a post or an user to be direct replies to that post or the last post authored by the mentioned user.
Intuitively, since Social Media provide dedicated tools to express positive feedback on a post, a reply is likely to include some degree of criticism.
This definition of attack is clearly a simplification of reality, however we claim that this rough approximation allows to extract meaningful information to the extents of recognising positive reception of content.

To enrich our modelling we are also considering supporting arguments.
For the sake of simplicity, we choose to consider as supporting arguments to a given post the messages that attack the replies to the considered post.
Since this notion of support is a simplified one and differs from its more established meaning, we will herein refer to it as \emph{defence}.

\subsection{Graph Construction Rules}
Building upon the considerations and assumptions illustrated in the previous paragraphs, we define two rules to construct a graph model from a conversation extracted from a social media platform:
\begin{itemize}
\item \emph{Attack rule}: a post $a$ is attacking another post $b$ whenever $a \neq b$, the author of $a$ is not the author of $b$, and $a$ is either a reply to $b$ or it contains an explicit referral to a user $u$ and $b$ is the last post authored by such a user at the time $a$ is posted. If $a$ is not a reply or does not contain any reference to a user, we assume that $a$ is attacking the previous post in the conversation authored by a different user than the one who wrote $a$.
\item \emph{Defence rule}: a post $a$ is defending another post $c$ whenever there exist a post $b$ such that $a$ attacks $b$ and $b$ attacks $c$.
\end{itemize}
As a corollary of the above rules, our approach will not include self-attacking nodes, this choice is motivated by the fact that social media conversations tend to include sequences of posts authored by the same user which would be an obvious error to model as a series of self-attacks. Self-defencing edges, on the other hand are allowed since it appears reasonable that a user may support his or her own opinions.
On the other hand, the same user cannot be both attacked and defended within a single post, since if $a$ attacks $b$ and defends $c$, $b$ and $c$ cannot be authored by the same user. 
\begin{figure}[ht]
    \centering
    \includegraphics[width=0.75\textwidth]{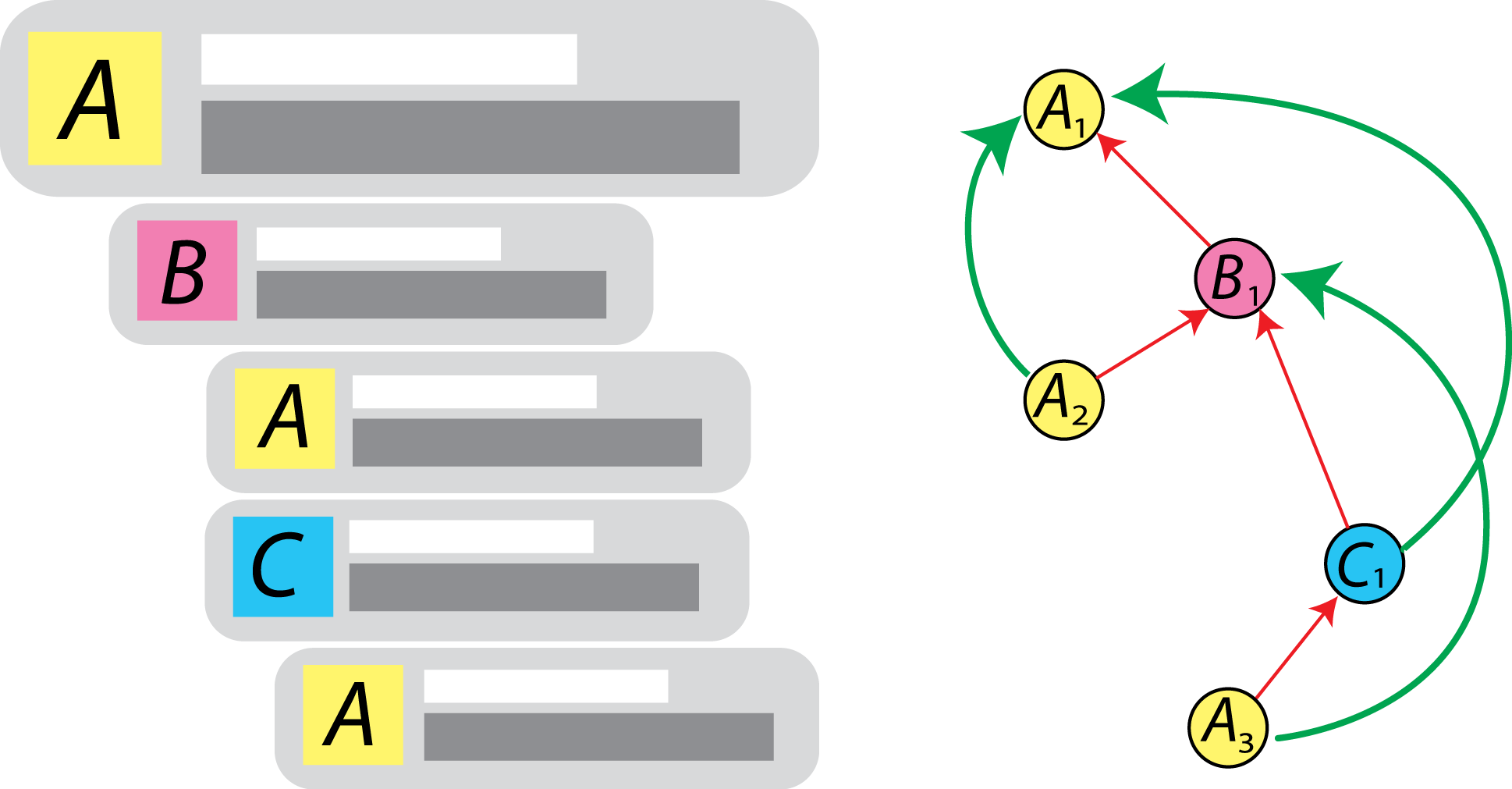}
    \caption{A fragment of an actual Reddit conversation and the graph it generates. Replies are modelled as attacks as long as the reply is not authored by the same user that wrote the original post, and defence edges are drawn between two posts a and c whenever there exist a post b such as a attacks b and b attacks c}
    \label{fig:aaf_build}
\end{figure}
In Figure \ref{fig:aaf_build} is presented an example of application of the above rules to a fragment of a conversation taken from Reddit\footnote{User names in the figure are purposely changed and some texts have been truncated to fit into the picture.}.
The above described rules are used in this work to build automatically graphs from conversation crawled from Social Media.

\section{Considered Features and Classification Algorithms}
\label{sec:consideredFeatures}
To identify the users who are likely to attract most positive feedback, we leverage the network structure provided by the graph obtained with our approach to extract information about the interactions between users and use such information to train a classifier.
Different combinations of classifiers and graph features have been experimented to assess how effectively highly appreciated users can be identified by relying solely on the structure of reply chains between social media posts.
In the first part of the section we are describing the considered graph and node features, while in the second part we are briefly introducing the various classification techniques tested.

\subsection{Considered Features}
\label{subsec:considered_features}
For each user involved in a conversation, we extracted a set of features representing his o her involvement and the reactions his or her posts generated.
We considered a broad set of features, shown in Table \ref{tab:propriotutto}, to model several different aspects of the obtained graphs.
We included among the considered features some basic distributional information, such as the number of posts authored by the considered user within the considered conversation, called \emph{Post Count}. We can normalise the Post Count with respect to the total number of posts occurring in the considered conversation, obtaining the \emph{Conversation Coverage} feature.

Moving onto the graph structure, we outline four features describing a user's observed activity: \emph{Received Attacks} defined as the cumulative number of attacks received by the posts of the user, \emph{Performed Attacks} defined as the cumulative number of attacks performed by the user through his or her posts, \emph{Received Defences} defined as the cumulative number of defences received by the posts of the user, and \emph{Performed Defences} defined as the cumulative number of defences performed by the user through his or her posts.
The above four features can be averaged over the Post Count, resulting in another four features: \emph{Average Received Attacks}, \emph{Average Performed Attacks}, \emph{Average Received Defences}, and \emph{Average Performed Defences}.
Further features can be obtained by combining the above ones to quantify some user behavioural aspects. In particular we focus on three possible combinations we call \emph{Aggressiveness}, \emph{Community Disapproval Ratio}, and \emph{Engagement}.
We define the Aggressiveness as the Performed Attacks divided by Performed Defences; such a value is greater than 1 when the attacks performed by a user outnumber his or her defences, otherwise it ranges between 0 and 1.
We define the Community Disapproval Ratio as the Received Attacks divided by Received Defences; such value represents the proportion between attacking and supporting arguments to the user's post throughout the conversation and therefore can be considered a measure of how much his or her contributions were disputed, with a high value representing a majority of attacks, and a value close to zero a majority of supporting arguments.
Finally, we define the Engagement as the sum of the four edge-based features, assuming that users who both receive and perform a large number of attack and defences can be regarded as highly involved in the conversation. The Engagement feature, being a count over the edges in the graph, can be normalised over the total number of such edges, resulting in the \emph{Normalised Engagement}; this feature represents the fraction of edges in the graph that sprout from the user's posts or enter in it.
We can wrap up the information provided by the Engagement and the number of post authored by the user into an \emph{Activity Score} which is defined as the product of Engagement and Conversation Coverage. This feature rewards users who post a lot and participate a lot by means of both attacks and defences, while punishing users who detain few hi-impact posts or too many low-impact ones.

Finally, we included features intended to describe how central in the graph topology are the posts authored by a user. We use three centrality measures: between centrality, which is, given a node, the number of shortest paths in the graph between two nodes that pass through it, Eigenvector centrality, which represents the number of connections of a node to other highly central nodes in the graph\footnote{Google's famous PageRank algorithm is a variant of Eigenvector centrality}, and Closeness centrality, which is the sum of the lengths of the shortest paths between the considered nodes and all the other nodes in the graph.
Centrality is a key concept of graph theory and can be used to estimate the overall relevance of a node in a network, thus it can embed information relevant to our purpose.
To represent this information, we introduce the last three features: \emph{Cumulative Between Centrality}, \emph{Cumulative Eigenvector Centrality}, and \emph{Cumulative Closeness Centrality}. Those features consist in the sum of the centralities in the network of all the posts authored by the considered user.


\begin{table}
    \resizebox{\textwidth}{!}{
        \begin{centering}
        \begin{tabular}{@{}llr@{}}
        \toprule
        \textbf{Feature Name}       & \textbf{Shorthand} & \textbf{Explanation}                                                \\ \midrule
Post Count                  & PC                 & Number of posts authored by the user                                \\
Conversation Coverage       & CC                 & PC/ posts in the conversation                                       \\ \midrule
Received Attacks            & Att-IN             & attack edges entering in the posts authored by the user             \\
Performed Attacks           & Att-OUT            & attack edges originating from the posts authored by the user        \\
Received Defences           & Def-IN             & defence edges entering in the posts authored by the user            \\
Performed Defences          & Def-OUT            & defence edges originating from the posts authored by the user       \\

Average Performed Attacks   & AvgAtt-OUT         & Att-OUT/PC                                                          \\
Average Received Attacks    & AvgAtt-IN          & Att-IN/PC                                                           \\
Average Performed Defences  & AvgDef-OUT         & Def-OUT/PC                                                          \\
Average Received Defences   & AvgDef-IN          & Def-IN/PC                                                           \\
Aggressiveness              & Agr                & Att-OUT/Def-OUT                                                     \\
Community Disapproval Ratio & Dis                & Att-IN/Def-IN                                                       \\
Engagement                  & En                 & Att-IN + Att-OUT + Def-IN + Def-OUT                                 \\
Normalised Engagement       & NEn                & En/ cumulative En of all users in the conversation                  \\
Activity score              & As                 & En * CC                                                             \\
Normalised Activity Score   & NAs                & As / max As in the conversation                                     \\
\midrule
Cumulative Betweenness  Centrality     & CBC                 & Cumulative Betweenness centrality of all posts authored by the user \\
Cumulative Eigenvector Centrality      & CEC                 & Cumulative Eigenvector centrality of all posts authored by the user \\
Cumulative Closeness Centrality        & CClC                & Cumulative Closeness centrality of all posts authored by the user  
        \end{tabular}
        \end{centering}
    }
    
\label{tab:propriotutto}
\caption{Considered user features.}
\end{table}

\subsection{Considered Classifiers}
\label{subsec:considered_classifiers}
Several classifiers have been considered to tackle the problem of identifying users whose content attract positive feedback. They are: Naive Bayes classifiers, Conditional Inference Trees, Random Forests, and Support Vector Machines.
A Naive Bayes classifier is a probabilistic classifier based on Bayes' theorem that leverages the strong independence assumption between the considered features. Such classifiers, albeit simple, are robust and cope well with hi-dimensional problems.
Moreover there is evidence in the literature that Naive Bayes classifiers perform well in classification tasks similar to the one presented in this work \cite{arnt2003learning}.
A Conditional Inference Tree is a classifier based upon rule induction, similar to a decision tree, built performing recursively univariate splits of the class value based on the values of a set of features. The splits in decision trees are learnt leveraging information metrics, such as the information entropy, while in a conditional inference tree they are chosen according to permutation-based significance tests, thus the selection bias of picking variables that have multiple splits is mitigated, reducing the risk of overfitting the training data \cite{hothorn2006unbiased}.
Random Forests are an ensemble learning technique consisting in building at training time several decision trees, with each of them based on a random subset of the training features, and then aggregating the various predictions to generate the final one \cite{breiman2001random}. This method mitigates the problem of training data overfitting to which decision trees are particularly prone.
Finally, Support Vector Machines (SVM) are the most established Kernel method algorithm, commonly used to solve a large number of classification and regression problems. While they can be considered as linear classifiers, their usage of kernel functions allows to map the training data points into a high dimensional space wherein the classification problem is linear, thus allowing non-linear classification. While very powerful, Support Vector Machines can suffer greatly biases induced by uncalibrated class membership probability in the training data.
In this study we are considering three variants of SVM, differing in the employed kernel function: the \emph{linear} SVM uses a linear kernel function which is the most common SVM setup for NLP tasks, the \emph{polynomial} SVM uses a polynomial kernel function of degree $d=2$, finally the \emph{radial} SVM uses Radial Basis Function kernel as kernel function, which provides in principle a low-band pass filter, selecting out smooth solutions.

\section{Data analysis and Evaluation}
\label{sec:dataAnalysisEval}
In this section we describe the evaluation work performed to assess the effectiveness of our approach which exploits abstract argumentation and machine learning in detecting users whose contributions are most appreciated by the community.
More specifically, we will describe the considered data source, the gathered data sets, the importance of the considered features, and the performance of the considered classifiers on the gathered data sets.

\subsection{Data Preparation} 
\label{sec:dataPrep}
To get meaningful real world data to perform both training and validation, we crawled conversations from Reddit\footnote{https://www.reddit.com}.
Reddit is an entertainment, social news networking, and news Web site whose community members can post messages or link to other Web pages, and reply to posts of other users. Each contribution can be voted by the community members with upvotes, expressing positive feedback, and downvotes, expressive negative feedback.
At the time of this writing, Reddit is among the 30 most viewed sites of the Web, receives over 1 billion visits each month\footnote{https://www.similarweb.com/website/reddit.com}, and in 2015 it saw 725.85 million submissions from 8.7 million total authors that received 6.89 billion upvotes\footnote{http://www.redditblog.com/2015/12/reddit-in-2015.html}.
One of the most notable characteristics of Reddit is the fact that it provides scoring mechanisms for both posts and users based on the upvotes and downvotes provided by the community over time.
In particular we are interested in the score associated to the single posts which is determined by the difference between upvotes and downvotes and can be seen as a straightforward approval rating of the considered content. On the other hand, the user score, called \emph{link karma}, is known for taking into account more than the explicit community feedback, but since its evaluation procedure is not clearly described by Reddit itself, it cannot be considered a reliable rating for our purpose.
Therefore we are estimating the general approval of a user's content by considering the cumulative scores of his or her posts (herein we will refer to such a score as \emph{cumulative approval}), without taking into account the link karma. It is important to point out how the score of a single post, being the difference between positive and negative community votes, may be a negative number, therefore a user whose post receive mixed feedback is expected to have a near zero cumulative approval.

We crawled the 125 longest conversations available on the social network, gathering posts from over 70,000 users. Graphs were constructed for each conversation according to the methodology described in Section \ref{sec:conversationModelling}, and for each user we evaluated the features described in Section \ref{sec:consideredFeatures}.
We observed the distribution of users with respect to their cumulative approval and we found that it respected the assumptions made in Section \ref{sec:intro}, with the vast majority of users concentrated around mildly positive values, and a few scoring extremely low or extremely high.
Considering the way the cumulative approval is computed and the shape of the distribution we can safely assume that the 5\% highest-scoring users can be regarded as the ones who produced the most appreciated content, herein we will refer to them as \emph{top users}.
The drawback of this choice is that it implies a strong unbalance between the considered classes, therefore to prevent statistical bias in the training of the classifiers we sampled the gathered data to obtain a set of 7,000 users equally split into top users and other users.
This balanced set, being a huge simplification of reality, is meant for training and preliminary evaluation purposes and we will refer to it in the rest of the paper as the \emph{Evaluation Set}.
To achieve a better representation of a real world scenario we crawled 100 more conversations picked among the longest ones hosted on the platform to create a new data set (herein called the Validation Set) unbiased and with a realistic distribution of users.
These conversations, like the ones in the previous set, were processed according to the methodology described in Section \ref{sec:conversationModelling} and \ref{sec:consideredFeatures}, then the 5\% top users were flagged according to their cumulative approval.
The produced validation set contains over 30,000 users, with 1622 of them flagged as top users.


\subsection{Exploratory Data Analysis}
\label{subsec:tiocà}
As a first step of our evaluation, we assessed which among the considered features are to be considered the most informative.
This analysis has two goals: determining if graph based features are actually informative and finding a minimal dimensionality for this problem, thus reducing the risk of the classifiers overfitting the data.

To determine whether or not the usage of graph derived features could benefit the identification of highly appreciated users, we considered three among the classifiers described in Section \ref{subsec:considered_classifiers}: Conditional Inference Tree, Random Forest, and Support Vector Machine with linear kernel.
We chose these three because each of them represents a class of classifiers: CIT is an explicit decision model, RF is and ensemble learning strategy, and SVM is a kernel method.
These classifiers were trained and tested by means of cross validation upon the Evaluation Set with three different feature sets: a set including no graph derived features, a set including also attack-derived features, and a set including all the 19 features described in Section \ref{subsec:considered_features}, thus including defence-derived features.
We assume that if a statistically significant difference is observable in all these three classifiers, the usage of graph features is to be considered significant as well.
Classifiers trained on the first set achieved the worst performance and classifiers trained on the full 19 ones the best performance; detailed results are shown in Table~\ref{tab:AAF_variants}. As expected, graph based features provided a considerable improvement, and in particular, the defence edges played an important role to further increase the results.

\begin{table}[]
\begin{tabular}{@{}llll@{}}
\toprule
Classifier              & Conditional Inference Tree & Random Forest & SVM    \\ \midrule
No graph                & 0.5376                     & 0.5648        & 0.5062 \\
Attacks                 & 0.8324                     & 0.8224        & 0.8257 \\
Attacks and Defences    & 0.8505                     & 0.8405        & 0.8338 \\ \bottomrule
\end{tabular}

\label{tab:AAF_variants}
\caption{Classification accuracy with no graph derived features, with attack derived features, and with all considered features.}
\end{table}

To identify the most informative features, an extensive exploratory data analysis was performed.
Four techniques were considered: Principal Component Analysis (PCA), Linear Discriminant Analysis (LDA), Learning Vector Quantization (LVQ), and   Recursive Feature Elimination (RFE).
PCA is a well-known technique to estimate a number of linearly independent components that represent most of the observed variability in the data, thus providing a better understanding of the minimum dimensionality that could be sufficient to cover the maximum variance.
LDA is another well known technique that takes into account information about objects classes, allowing us investigate which are the directions (called “linear discriminants”) that represent best the axis that maximise the separation between the ``top users'' and the rest of the user base.
Finally, LVQ and RFE leverage a predictive model, such as a Random Forest, to estimate how relevant the considered features are. More precisely, LVQ estimates features importance using a ROC curve analysis conducted for each feature, and RFE generates several sub-sets of the feature set, then it trains the model on such sub-sets, and evaluates the model's performance to identify the most significant among the generated feature sub-sets.

For the purposes of these analysis the first gathered data set (the one including 125 conversations) was considered.
Without further venturing into the details of the above described analysis, the insights gathered by means of PCA, LDA, and LVQ were coherent and highlighted the following features as the most informative of the considered set: Average Received Attacks (AvgAtt-IN), Average Received Defences (AvgDef-IN), Engagement (En), and Cumulative Betweenness Centrality (CBC). 
Notably, the Engagement value appeared to be consistently the strongest predictor throughout the performed feature importance analysis.
We will refer to these four features together as the \emph{minimal} feature set.
The RFE approach, instead suggested that the best performance on the considered data set could be achieved with a slightly broader feature set, including Received Defences (Def-IN), Conversation Coverage (CC), Average Performed Attacks (AvgAtt-OUT), Average Received Attacks (AvgAtt-IN), Average Received Defences (AvgDef-IN), Community Disapproval Ratio (Dis), and Engagement (En).
We will refer to these seven features as the \emph{reduced} feature set, in contrast with the \emph{full} feature set including all the 19 considered features.

\subsection{Classification Experiments}
\label{sec:experiments}
We ran two experiments on the two gathered data sets to find the most suitable combination of feature set and classification model. The first experiment consisted in performing a cross-fold evaluation over the Evaluation Set with all the classifiers presented in Section \ref{subsec:considered_classifiers} using the full feature set.
The results of this evaluation are shown in Table \ref{tab:considered_class_acc} where, for each considered classifier, its precision ranges are shown. Each range has a lower and an upper bound that enclose all the observed values with 0.95 probability, and the average observed value.
\begin{table}[]
    \resizebox{\textwidth}{!}{
\centering
\begin{tabular}{@{}lllllll@{}}
\toprule
Accuracy & Naive Bayes & Random Forest & Conditional Inference Tree & SVM Linear & SVM Poly & SVM Radial \\
\midrule
lower bound  & 64.36\%  & 83.90\%       & 83.45\%                    & 82.62\%    & 81.58\%  & 79.74\%    \\
average      & 66.43\%  & 84.05\%       & 85.05\%                    & 83.38\%    & 83.52\%  & 82.19\%    \\
upper bound  & 68.45\%  & 84.10\%       & 86.55\%                    & 84.87\%    & 84.94\%  & 82.85\%    \\ \bottomrule
\end{tabular}
}
\label{tab:considered_class_acc}
\caption{Considered classifiers accuracy range with full feature set measured upon cross-validation.}
\end{table}
\begin{table}
\resizebox{\textwidth}{!}{
\centering
\begin{tabular}{@{}llllll@{}}
\toprule
Metric    & Random Forest & Conditional Inference Tree & SVM Linear & SVM Poly  & SVM Radial \\ \midrule
Precision & 0.8762        & 0.7867                     & 0.8943     & 0.8657    & 0.8838     \\
Recall    & 0.8156        & 0.9017                     & 0.7877     & 0.8160    & 0.7864     \\
F1        & 0.8448        & 0.8403                     & 0.8433     & 0.8401    & 0.8323     \\ \bottomrule
\end{tabular}

}

\label{tab:good_result}
\caption{Precision, Recall and F1 score of the considered classifiers measure upon cross-validation.}
\end{table}
It can be easily noticed how the Naive Bayes classifier achieves a significantly lower precision (between 64.36\% and 68.45\%) than the other considered models that are relatively tied with scores between 80\% and 86\% precision. This observation led us to the withdrawal of the Naive Bayes classifier from the considered classifiers pool.
Moreover, since what we are really interested in the retrieval of appreciated users, we considered some Information Retrieval metrics, namely Precision, Recall, and their harmonic mean the F1 score, all evaluated on the ``top user'' class.
The values of such metrics achieved by the five best-performing classifiers is shown in Table \ref{tab:good_result}.
All classifiers were trained with the full feature set, and Random Forests and Conditional Inference Trees appear to achieve the most promising results, performing consistently better than SVM in various configurations.
These results can be improved up to 2\% average classification accuracy by adopting the reduced feature set instead of the full one, however, being such a feature set selected with RFE on this very same data (see Section \ref{subsec:tiocà}) such an improvement is expected and might be the outcome of some data overfitting.
%
%
%
The second experiment consisted in training the classifiers on the whole Evaluation Set and then run them over the Validation Set, which presents a more realistic distribution of users than the Evaluation Set. 
The evaluation was performed over the three chosen feature sets to identify the one that describes best the problem with respect to a possible field usage.
The results of this second evaluation are presented in Table~\ref{tab:classifiers_acc}, where the observed precision values achieved by each combination of feature set and classifier is shown.
However, due to the strong unbalance of classes in the Validation Set, with 94.9\% of users not being top users, classification accuracy alone is little informative.
To cope with this characteristic of the domain, the aforementioned Precision, Recall, and F1 score, evaluated on the ``top user'' class, are more informative and their values are shown in Table \ref{tab:prf1_minimal}, \ref{tab:prf1_reduced}, and \ref{tab:prf1_full}.

\begin{table}{\resizebox{\textwidth}{!}{
\centering
\begin{tabular}{@{}llllll@{}}
\toprule
Feature Set & Random Forest & Conditional Inference Tree & SVM Linear & SVM Poly & SVM Radial  \\ \midrule
minimal     & 90.24\%       & \textbf{90.96\%}                    & 87.68\%    & 88.93\%   & 88.00\%       \\
reduced     & 86.84\%       & 90.75\%                    & 88.04\%    & 75.04\%   & 80.33\%       \\
full        & 83.95\%       & 90.85\%                    & 79.78\%    & 81.01\%   & 77.66\%           \\ \bottomrule
\end{tabular}
}
}
\caption{Considered classifiers average accuracy on Validation Set.}
\label{tab:classifiers_acc}
\end{table}

\begin{table}{\resizebox{\textwidth}{!}{
\centering
\begin{tabular}{@{}llllll@{}}
\toprule
Metric    & Random Forest & Conditional Inference Tree & SVM Linear & SVM Poly & SVM Radial  \\ \midrule
Precision & 0.3155        & 0.3332            & 0.2651     & 0.2865    & 0.2696     \\
Recall    & 0.8021        & 0.7947                     & 0.8163     & 0.8046    & 0.8095     \\
F1        & 0.4529        & 0.4695                     & 0.4002     & 0.4225    & 0.4045     \\ \bottomrule
\end{tabular}
}
}
\caption{Precision, Recall and F1 score of the considered classifiers with the minimal set of features.}
\label{tab:prf1_minimal}
\end{table}

\begin{table}{\resizebox{\textwidth}{!}{
\centering
\begin{tabular}{@{}llllll@{}}
\toprule
Metric    & Random Forest & Conditional Inference Tree & SVM Linear & SVM Poly & SVM Radial  \\ \midrule
Precision & 0.2557        & 0.3290                     & 0.2711     & 0.1587    & 0.1871     \\
Recall    & 0.8452        & 0.8058                     & 0.8144     & 0.9199 & 0.8693     \\
F1        & 0.3927        & 0.4673                     & 0.4068     & 0.2706    & 0.3079     \\ \bottomrule
\end{tabular}
}
}
\caption{Precision, Recall and F1 score of the considered classifiers with the reduced set of features.}
\label{tab:prf1_reduced}
\end{table}

\begin{table}{\resizebox{\textwidth}{!}{
\centering
\begin{tabular}{@{}llllll@{}}
\toprule
Metric    & Random Forest & Conditional Inference Tree & SVM Linear & SVM Poly & SVM Radial  \\ \midrule
Precision & 0.2231        & 0.3318                     & 0.1872     & 0.1961    & 0.1701     \\
Recall    & 0.8816        & 0.8064                     & 0.9026     & 0.8940    & 0.8859     \\
F1        & 0.3561        & 0.4702                     & 0.3101     & 0.3216    & 0.2853     \\ \bottomrule
\end{tabular}
}
}
\caption{Precision, Recall and F1 score of the considered classifiers with the full set of features.}
\label{tab:prf1_full}
\end{table}


\section{Discussion}
\label{sec:discussion}
The insights provided by our evaluation activities described in Section \ref{sec:dataAnalysisEval} appear to substantially uphold the claim introduced in Section~\ref{sec:intro}, highlighting how the structure of the Social Media conversations embed information about the posts appreciation level, and hence posts' author reputation.

First of all, the distribution of the Reddit score, as noted in Section~\ref{sec:dataPrep} and according to what we expected, our assumption that only a small fraction of the user base is able to consistently generate well received content.
With respect to such a distribution our choice of considering ``top users'' the highest scoring 5\% of the user base seems reasonable, given that such fraction is likely to include the whole long tail of highly appreciated users.
Before discussing the outcomes of the evaluation it is important to stress how the Reddit score, though being strongly correlated with the number of posts, represents a rather qualitative dimension of a user's activity in the community.
As a matter of fact users who post a lot tend to accumulate more Reddit Score than users who seldom post, however, due to the possibility of expressing negative feedback, the correlation between the post count and the RS does not go beyond 0.7, a significant, but not outstanding value. This is particularly striking when comparing Reddit to other social networks such as Facebook wherein the amount of ``likes'' received by a user appears to be close to being a function of the number of posts authored~\cite{bessi2015everyday} with little regard for the actual content posted.
Considering, as we did, the average Reddit Score, i.e. the average proportion of upvotes and downvotes received by a user's post, moves even forward the qualitative aspect of this score.
The usage of AAF to model conversations, despite the numerous simplifying modelling assumptions introduced in Section~\ref{sec:conversationModelling}, appears to be instrumental to the extents of our task.
The results shown in Table~\ref{tab:AAF_variants} suggest that without considering information derived from our graph-based modelling, the precision of determining whether or not a user should be considered a ``top'' one is extremely low and dangerously close to a ``coin flip'' regardless of the employed classifier.
Attacks edges alone can provide significant information, but defences are informative as well and their addition appears to benefit the overall accuracy disregarding the employed classifier.

Since graph derived features appear to be informative, we tried to understand the performance limits of this technique by experimenting on different data sets with different combinations of feature sets and classifiers.
The first experiment described in \ref{sec:experiments} consisted in a cross evaluation over a balanced dataset, which means a very controlled scenario meant to avoid introducing statistical bias into the considered classifiers.
Moreover, the considered dataset was built crawling Reddit's most popular discussions, which means extremely large ones.
Keeping this in mind, the results gathered are nevertheless very encouraging with a high classification accuracy (peaking between 83.45\% and 86.55\%) and with five classifiers tied around a 0.84 F1 score on the ``top user'' class.
These results allow us to assume that in such a scenario our approach could achieve satisfying performance even employing different classifiers.

Reality, however, proved to not be so bright when the classifiers trained on the full Evaluation set were benchmarked over the more various Validation Set. The Validation Set most notable features are: inclusion of smaller and less popular discussions, way larger number of considered users, and a realistic proportion between highly appreciated users and mildly or not appreciated ones.
In this scenario our approach achieved an higher performance, scoring at best 90.96\% accuracy when considering only the minimal set of features which appears to be less prone to overfitting.
On the other hand, the thin numbers of highly appreciated users imply that a naive classifier opting always for the negative class would score an extremely high accuracy, scoring 94.9\%.
However, in such an unbalanced scenario, classification accuracy alone cannot be considered truly informative, especially when the goal is to retrieve only the members of one class.
Precision, Recall and F1 score, then, provide a better picture of the situation. The last one, in particular, allows us to assess the robustness of our classifiers being the harmonic mean of Precision and Recall; with respect to this metric, a naive classifier returning always the negative class would score zero, because its recall would be zero. Another naive classifier returning the positive class (top user) would score instead 0.09, having a 100\% recall, but a very low precision.

The five classifiers that in the previous experiment were tied in terms of F1 score, in this second trial appear to achieve significantly different results. SVMs in particular seem to perform consistently worse than Conditional Inference Trees and Random Forests.
All classifiers, except for Conditional Inference Trees, achieve a better performance on the minimal feature set, providing evidence that the larger feature sets give in to overfitting.
With respect to the results displayed in Table \ref{tab:good_result}, the ones in Table \ref{tab:prf1_minimal}, \ref{tab:prf1_reduced}, and \ref{tab:prf1_full} show a significantly lower Precision, but retain a similar Recall, implying a very thin number of false negative ``top users''.
Overall, Conditional Inference Trees appear to be the most robust kind of classifier for this task, consistently achieving best or near-best scores and showing little performance variation over different feature sets.
As a final note, the percentage of users flagged as ``top users'' by the five considered classifiers ranged between 12\% and 9\%, implying that, though being somewhat ``optimistic'' the considered classifiers come rather close to the actual split in the data.

\section{Conclusions}
\label{sec:conclusions}
In this work we presented a novel approach towards the detection of users who create, from their community's point of view, quality content. With respect to related work which focus on the detection of malicious or unappreciated behaviour, we shifted towards the ``best'' part of the user base that recent literature suggests to be more influential and therefore interesting for trend detection tasks and thus for opinion mining and marketing purpose.
With respect to this problem, our main contribution is in the content agnostic nature of the proposed approach, which by solely observing the flow of a discussion, i.e. the sequence of posts and replies, can identify users who are likely to attract the support of their online community.
This approach, in fact, allowed us to generate reasonably small sets of potentially highly appreciated users that included up to 90\% of the actual ``top users''.
Unlike widespread content-based Sentiment Analysis techniques that leverage the content of the posts to detect positive attitudes, our approach abstracts over several non trivial problems such as irony detection by looking at the whole structure of the discourse.
With respect to argument mining, our approach does not investigate the content of messages included in the conversation, but rather their sequence and the order in which users reply to each other, suggesting that this sole information can embed insights about users' attitude.
The results gathered so far suggest that tackling the problem at discourse level allows to significantly restrict the search space for this problem, retaining a very high recall.

Our future work will be aimed at further exploring the possibilities of applying AAF to social media mining.
In particular, we will investigate the usage of the approach described in this paper as a preliminary step in more complex analysis, such as the detection of opinion leaders and other kinds of social influencers.
We will also experiment extensions to the approach herein described, with different definitions of defences and attacks.
Our discourse-based approach will also be coupled with more traditional message content-based Sentiment Analysis techniques to overcome the limitations of both techniques and to further investigate the relationships among social media users.
Finally, we believe that this work suggests a deeper connection between formal Artificial Intelligence techniques and statistical ones.
More specifically, we believe that the former ones could be used to achieve a better modelling of the data, thus extracting more abstract, meaningful, and understandable features that can substantially help describing complex phenomena like social media interaction.


\bibliographystyle{elsarticle-harv} 
\bibliography{biblio.bib}

\end{document}